%% file: manuscript.tex
\documentclass{article}

\usepackage{amssymb}
\usepackage{spconf,amsmath,graphicx}
\usepackage{multirow}
\usepackage{enumitem}
\usepackage{nicematrix}
\setlist{nosep, leftmargin=14pt}
\usepackage{lipsum} 
\usepackage{paralist}
\usepackage{mwe} 
\usepackage{enumitem}
\usepackage{graphicx}
\usepackage{subcaption}
\usepackage[linesnumbered,ruled,vlined]{algorithm2e}
\usepackage{stfloats}  
\usepackage[capitalize]{cleveref}

\crefname{section}{Sec.}{Secs.}
\Crefname{section}{Section}{Sections}
\Crefname{table}{Table}{Tables}
\crefname{table}{Tab.}{Tabs.}

\usepackage{graphicx}
\usepackage{array}
\usepackage{booktabs}
\usepackage{adjustbox}
\usepackage{xcolor} 
\usepackage{pgfmath} 

\newcommand{\sigstar}{\textsuperscript{\tiny *}}

\title{Enhancing Uncertainty Estimation in Semantic Segmentation via Monte-Carlo Frequency Dropout}

\name{
  Tal Zeevi$^1$,
  Lawrence H. Staib$^{1,2,3,*}$,
  John~A.~Onofrey$^{1,2,4,*}$
}

\address{
    Departments of 
    $^1$Biomedical Engineering, 
    $^2$Radiology \& Biomedical Imaging,  
    $^3$Electrical Engineering, \\
    $^4$Urology, 
     Yale University, New Haven, CT, USA 
    }

\begin{document}

\maketitle

\begin{abstract}
Monte-Carlo (MC) Dropout provides a practical solution for estimating  predictive distributions in deterministic neural networks. Traditional dropout, applied within the signal space, may fail to account for frequency-related noise common in medical imaging, leading to biased predictive estimates. A novel approach extends Dropout to the frequency domain, allowing stochastic attenuation of signal frequencies during inference. This creates diverse global textural variations in feature maps while preserving structural integrity -- a factor we hypothesize and empirically show is contributing to accurately estimating uncertainties in semantic segmentation. We evaluated traditional MC-Dropout and the MC-frequency Dropout in three segmentation tasks involving different imaging modalities: (i) prostate zones in biparametric MRI, (ii) liver tumors in contrast-enhanced CT, and (iii) lungs in chest X-ray scans. Our results show that MC-Frequency Dropout improves calibration, convergence, and semantic uncertainty, thereby improving prediction scrutiny, boundary delineation, and has the potential to enhance medical decision-making.

\end{abstract}
\begin{keywords}
Segmentation Uncertainty, Monte-Carlo Dropout, Frequency Dropout, Selective Prediction
\end{keywords}

\section{Introduction}
\label{sec:intro}
  \let\thefootnote\relax\footnotetext{ \text{*} These authors jointly supervised this work.}

    \let\thefootnote\relax\footnotetext{ \text{©} 2025 IEEE. Personal use of this material is permitted. Permission from IEEE must be obtained for all other uses, in any current or future media, including reprinting/republishing this material for advertising or promotional purposes, creating new collective works, for resale or redistribution to servers or lists, or reuse of any copyrighted component of this work in other works.}
  

Estimating prediction uncertainties in deterministic deep learning models
often involves the strategic introduction of controlled artificial noise into the data ~\cite{huang2024review}. This can occur either before~\cite{wang2019aleatoric, wang2019automatic} or during~\cite{karimi2019accurate, liu2020exploring, nair2020exploring} neural network processing, with subsequent measurement of variations in model performance to assess robustness.  Techniques such as DropConnect~\cite{wan2013regularization} and Dropout~\cite{srivastava2014dropout}, which randomly omit network edges or nodes during processing, have been foundational in this respect, effectively injecting random patterns of noise into the network's operation allowing the simulation of a predictive distribution approximating Bayesian inference~\cite{gal2016dropout}.

In convolutional neural network (CNN) layers, commonly used in segmentation tasks, each convolution step corresponds to a node on the network graph, essentially turning Dropout into a random source of impulse noise within the CNN feature maps. This method, however, may not comprehensively capture the predictive distribution in medical imaging, where noise extends into the frequency domain -- a range poorly addressed by impulse noise. Our recent findings~\cite{zeevi2024monte} suggest that Frequency Dropout ~\cite{khan2019regularization}, which randomly removes frequency components from feature maps during Monte Carlo (MC) simulations, refines predictive uncertainty estimates in medical imaging classification  over traditional Dropout. 

\input{figures/figure1}

In semantic segmentation, preserving precise structural integrity is essential, sometimes down to the pixel level. Impulse noise introduced by traditional Dropout can significantly impact spatial features, causing non-uniform and abrupt changes in the gradient field. This can disrupt fine edges, distort subtle imaging markers, or introduce high variability in otherwise uniform regions, as exemplified in ~\cref{fig1}. Traditional Dropout operates independently on individual elements of the feature map, without accounting for dependencies across spatial features.  In contrast, Frequency Dropout's attenuations in the frequency domain generate a global noise effect on the feature maps, which may better preserve structural relationships and dependencies between spatial features. This approach may more accurately capture uncertainty, particularly in regions where preserving feature correlations, such as object boundaries, is essential.

In this study, we explore MC-Frequency Dropout for semantic segmentation to assess its impact on uncertainty estimation. Our contributions are: (i) defining Signal and Frequency Dropout within CNN layers, highlighting their theoretical foundations and effects on structural integrity; (ii) comparing uncertainty estimates across diverse modalities, including X-ray, CT, and MRI; and (iii) examining how the placement of dropout layers within state-of-the-art medical imaging networks influences segmentation performance, demonstrating the benefits of strategic dropout positioning.

\section{Background}
\label{sec:background}

Without loss of generality, let $C_\theta : \mathbb{R}^{m \times n} \rightarrow \mathbb{R}^{m' \times n'}$ denote a convolution block within a CNN model $M_\Theta$, comprising a single convolutional layer, where $\theta \in \Theta$ represents the set of trainable model parameters. The forward pass output of $C_{\theta}$ for a given input instance $X\in \mathbb{R}^{m\times n}$ is expressed as follows:
\begin{equation*}
C_{\theta}(X) = \sigma(X \ast W + b) 
\end{equation*}

where $W^{k \times k} \in \theta$ represents the convolution kernel, $b \in \theta$ denotes the bias term, $\ast$ signifies the convolution operation between $X$ and $W$, and $\sigma(\cdot)$ denotes an element-wise non-linearity (e.g., rectified linear unit activation function). 

\subsection{Signal Diluted Forward Pass} In a signal-diluted forward pass \cite{srivastava2014dropout}, the convolution kernel is multiplied by a binary mask in each convolution step: 
\vspace{-3pt}
$$(X \ast W)(i, j) =$$
\vspace{-35pt}

\small\begin{flalign*}
\sum_{u=1}^{k} \sum_{v=1}^{k} X(i + u - \lfloor k/2 \rfloor, j + v - \lfloor k/2 \rfloor)
&\cdot W(u, v) \cdot D_{ij}(u,v).
\end{flalign*}

\normalsize

\noindent In this equation, $i$, $j$ index the output feature map, $u$, $v$ index the kernel, and $D_{ij}^{k \times k}$ acts as a binary dropout mask, where each element $D_{ij}(u,v)$ is independently sampled from a Bernoulli distribution with probability $p$.
 
Dropout is the operation of selectively disabling nodes from the network graph during the forward-pass. In convolutional layers, this process involves setting all kernel weights to zero during a particular convolution step, denoted by $D_{ij}(u,v)=D_{ij}(u',v'), ~\forall u,v,u',v'$. This action separates the dropout process from the convolution operation itself. As a result, the output from a CNN block that includes Signal Dropout, labeled as $C_{\theta}^D$, is expressed as:

\begin{equation*}
C_{\theta}^D(X) = \sigma([X \ast W]\odot D + b).
\end{equation*}

Here, $D^{m'\times n'}=[d{(i,j)}]$ acts as a binary dropout mask and the operator $\odot$ represents element-wise multiplication. 

Applying the mask \(D\) via element-wise multiplication involves \(m' n'\) operations, resulting in a computational complexity of \(O(m' n')\) for Signal Dropout.

\subsection{Frequency Diluted Forward Pass}

The frequency-diluted forward pass \cite{zeevi2024monte} removes signal frequencies within feature maps. Instead of applying the dropout mask $D$ directly to the signal, we transform the signal to the Fourier space, apply the mask to remove frequency components, and reconstruct the signal using the inverse Fourier transform. The output of a forward pass through a CNN block with frequency dilution, denoted as $C_{\theta}^\mathcal{F}$, is expressed as: 

\begin{equation*}
    C_{\theta}^\mathcal{F}(X)= \sigma(\mathcal{F}^{-1}(\mathcal{F}[X \ast W] \odot D) +b).
\end{equation*}

Here, $\mathcal{F}$ and $\mathcal{F}^{-1}$ represent the Fourier transform and its inverse, respectively.

Using the Fast Fourier Transform (FFT), each transformation has complexity of \(O(N \log N)\), where \(N = m' n'\) is the number of elements in the feature map. Thus, the overall computational complexity of Frequency Dropout is \(O(m' n' \log(m' n') + m' n')\), accounting for FFT operations and element-wise multiplication with the dropout mask.

\section{Empirical Evaluation}
\label{sec:experiments}
We compared uncertainty estimates from Monte Carlo (MC) simulations with Signal and Frequency Dropout during inference to identify segmentation errors across three public semantic segmentation tasks involving MRI, CT, and X-ray modalities. The CT and MRI datasets were sourced from the Medical Segmentation Decathlon \cite{antonelli2022medical}
. The source code for our proposed method is publicly available$^{1}$.
\let\thefootnote\relax\footnotetext{$^{1}$ 
https://github.com/talze/frequency-dropout.git}

\subsection{Segmentation Datasets}

\paragraph*{Prostate zones on biparametric MRI (bpMRI) scans:} 
36 transverse T2-weighted and apparent diffusion coefficient (ADC) MRI scans of the prostate, each annotated to delineate two adjoining prostate zones: the peripheral zone (PZ) and the transitional zone (TZ) \cite{antonelli2022medical}.

\paragraph*{Liver and tumors on contrast enhanced CT scans:}
37 contrast-enhanced liver CT scans with liver and tumor annotations from patients with metastatic liver disease. \cite{antonelli2022medical}.

\paragraph*{Lungs on Chest X-ray scans:}
40 annotated chest X-rays from the National Library of Medicine's digital image database for Tuberculosis (TB) \cite{candemir2013lung, jaeger2013automatic}, featuring TB and normal cases, with the lung area behind the heart excluded.

\subsection{Segmentation Models}
For the prostate MRI and liver CT segmentation tasks, we used the top-performing pre-trained models available in the Medical Segmentation Decathlon \cite{antonelli2022medical}. In both tasks, these models were developed using the nnU-Net approach, which implements a self-configuring segmentation pipeline based on the U-Net architecture \cite{isensee2021nnu}. For the lung X-ray task, we utilized the MedSAM transformer model \cite{ma2024segment}, adapted from the Segment Anything (SAM) model \cite{kirillov2023segment}. This model is designed for broad medical image segmentation and is applicable to a range of modalities, including X-ray.

\subsection{Dropout Variants in Segmentation Networks}
We devised three model dropout variants, each with distinct dropout layer placements: \begin{inparaenum}[(i)]
\item Encoder Dilution: Dropout layers are introduced after each encoder step, with no dilution in the decoder section.
\item Decoder Dilution: Dropout layers are introduced after every decoding step, except for the output step, with no dilution in the encoder section.
\item Global Dilution: Dropout layers are introduced after every encoding or decoding step, except for the output step.
\end{inparaenum}

\subsection{Monte-Carlo (MC) Dropout Simulation}
For each input instance, we performed $R=30$ forward-pass repetitions, each with a new random diluted realization of the pre-trained U-Net or SAM models, using two dropout strategies: (i) Signal Dropout and (ii) Frequency Dropout. 
Voxel-level prediction estimates were obtained from the arithmetic mean of SoftMax values across Monte Carlo repetitions, with predictive uncertainty estimated from their standard deviation \cite{gal2016dropout}. This procedure was repeated across various dropout rates ($p$=0.01, 0.02, 0.04, 0.08, 0.16, and 0.32) to evaluate the sensitivity of the dropout approaches to changes in dropout rate.

\subsection{Evaluation Metrics}
\paragraph*{Expected Uncertainty Calibration Error (UCE):} 
Uncertainty estimates can be used to reject uncertain predictions, acting as scores to classify predictions into binary outcomes: reject or do-not-reject, a process known as selective prediction \cite{mukhoti2018evaluating, mobiny2021dropconnect}. We evaluated the alignment of MC-Dropout uncertainty estimates with voxel-level segmentation errors from the full (no-dropout) model using UCE \cite{laves2019well}. UCE, similar to Expected Calibration Error (ECE) \cite{guo2017calibration}, measures the calibration of uncertainty scores rather than 
predicted probabilities.

\paragraph*{Dice Similarity Coefficient (DSC):} Diluting neural network models can impact performance. We measured the deviation in segmentation accuracy between the diluted models and the full (no-dropout) model using the Dice Similarity 
Coefficient \cite{antonelli2022medical}. 

\noindent
Both UCE and DSC were calculated per instance and averaged across the cohort to assess overall model performance.

\section{Results}

\begin{figure*}[tb!]  
    \centering

        \includegraphics[width=\textwidth]{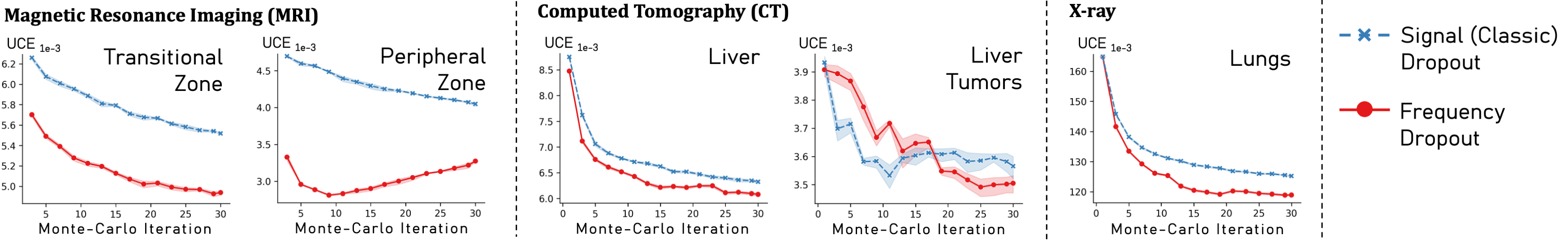}
    
   \caption{\small Calibration of uncertainty estimates for the best dropout configurations (rates and layer placements) of frequency (red) and signal dropout (blue). Subplots show the discrepancy between MC-dropout uncertainties and full (no-dropout) model errors. Frequency dropout aligns uncertainty estimates more accurately across tasks,  while preserving segmentation performance similar to the full model (Table \ref{table1}).}
    \label{best_ece}
\end{figure*}

\begin{table*}[h!]
\centering
\renewcommand{\arraystretch}{0.9} 
\setlength{\tabcolsep}{6.5pt} 
\small 
\caption{Impact of dilution on segmentation performance, reflected by DSC divergence between the dropout and full models}
\begin{adjustbox}{max width=\textwidth}
\begin{tabular}{@{}llcrrrrr@{}}
\toprule
 \textbf{Task} &  \textbf{Modality} & \textbf{DSC at baseline} & \multicolumn{4}{c}{\textbf{$\%$ Divergence from baseline DSC after $R$ Monte-Carlo iterations (sd)}} \\
\cmidrule(lr){3-3} \cmidrule(lr){4-7}
 &  & 
& \multicolumn{2}{c}{\textbf{$R$=5}} & \multicolumn{2}{c}{\textbf{$R$=30}} \\
\cmidrule(lr){4-5} \cmidrule(lr){6-7}
 &  & \textbf{No} {\footnotesize Dropout}   & \textbf{Signal} {\footnotesize Dropout} & \textbf{Frequency} {\footnotesize Dropout} & \textbf{Signal} {\footnotesize Dropout} & \textbf{Frequency} {\footnotesize Dropout} \\
\midrule
Liver               & CT   & 0.915 &  0.24 (0.31)  &  0.40 (0.41)  &  0.26 (0.36)  &  0.39 (0.38)  \\
Liver Tumors        & CT   & 0.606 & \sigstar\textcolor{red!70!black}{-14.53 (1.12)} & 1.23 (1.05)  & \sigstar\textcolor{red!70!black}{-11.72 (1.03)} & \sigstar\textcolor{blue}{3.54 (0.99)}  \\
Lungs               & X-ray & 0.831 &  0.44 (0.14)  & \sigstar\textcolor{blue}{0.91 (0.12)}  &  0.63 (0.11)  & \sigstar\textcolor{blue}{1.05 (0.10)}  \\
Transitional Zone   & MRI  & 0.781 & -0.05 (0.59)  & -0.11 (0.50) & -0.03 (0.57)  & -0.14 (0.42) \\
Peripheral Zone     & MRI  & 0.627 & -0.03 (0.58)  & -0.98 (0.46) & -0.12 (0.58)  & -1.02 (0.52) \\
\bottomrule
\end{tabular}
\end{adjustbox}

\vspace{2pt} 

\noindent\begin{minipage}{\textwidth}
\footnotesize{* Statistically significantly different from the baseline model ($p < 0.05$, Bonferroni corrected).}
\end{minipage}

\label{table1}
\end{table*}

Across all segmentation tasks, the best-performing Frequency Dropout configuration converged to better-calibrated uncertainty estimates than Signal Dropout, with uncertainties more closely matching full model segmentation errors (\cref{best_ece}). Similar performance was observed for liver tumor segmentation. Frequency Dropout achieved higher or comparable DSC scores relative to Signal Dropout and the full (no-dropout) model, demonstrating stable segmentation performance across tasks (\cref{table1}).

\begin{figure*}[tb!]  
    \centering
\includegraphics[width=\textwidth]{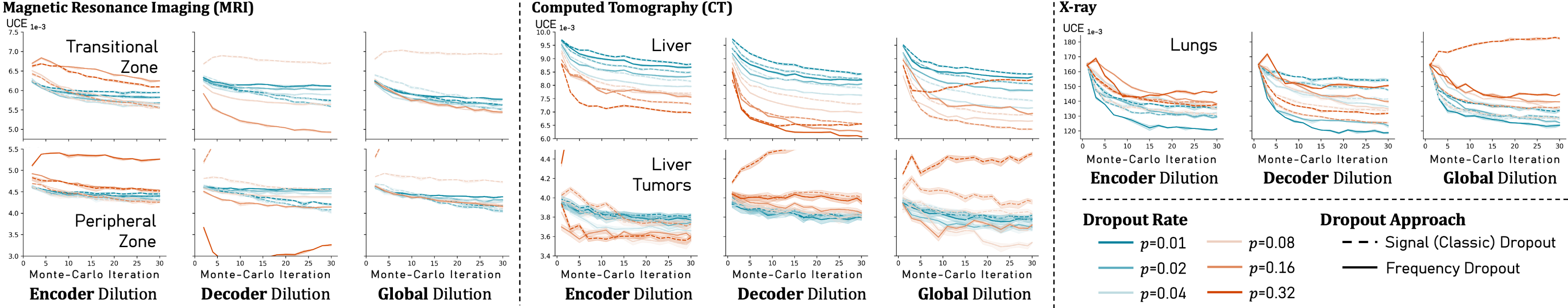}
    
    \caption{\small Calibration of MC-dropout uncertainty estimates with model errors, measured by Expected Calibration Error (ECE) ($\downarrow$) for different dropout configurations, including dropout rates and layer placements: Encoder Dilution, Decoder Dilution, and Global Dilution.}
    
    \label{rates}
\end{figure*}

\begin{figure*}[tb!]
\includegraphics[width = \textwidth]{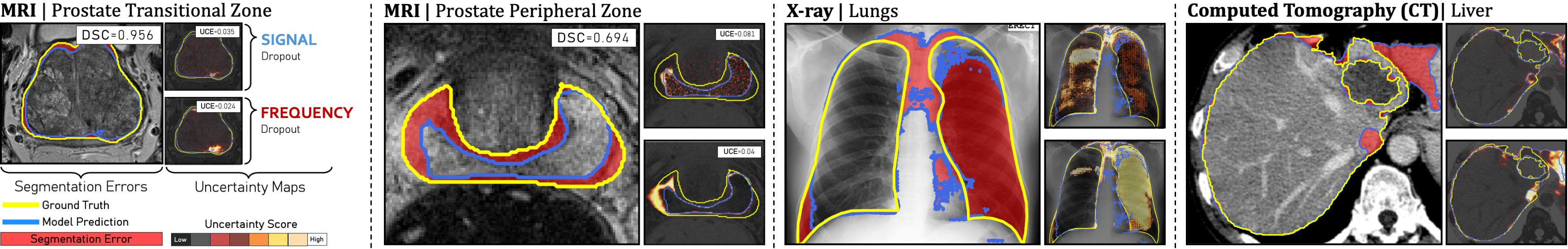}

\caption{\small MC-dropout uncertainty maps for segmentation tasks using Signal and Frequency Dropout. Large subplots show regions of ground truth (yellow line), full model prediction (blue line), and segmentation errors (red region). Small subplots present uncertainty maps: top for Signal Dropout, bottom for Frequency Dropout, with uncertainty measured as the standard deviation of voxel-level predictions across MC repetitions. 
Each plot uses the optimal dropout configuration for its approach.}
\vspace{-12pt}
\label{ue_map}
\end{figure*}

Performance varied across different dropout layer placements (\cref{rates}). In liver and most prostate segmentation tasks, Signal Dropout performed better in the encoder and global settings, while Frequency Dropout excelled in the decoder setting. For liver tumors, higher dropout rates were more effective with encoder dilution, while lower rates worked better with decoder dilution. In lung segmentation, Frequency Dropout maintained stable performance across all placements. Lower dropout rates generally improved the calibration of uncertainty estimates, though not always optimally, while higher rates introduced variability depending on placement but excelled in certain tasks. Specifically, higher rates improved liver segmentation, while lower rates were preferable for lung segmentation. In prostate segmentation, lower rates showed minimal variation, whereas higher rates led to significant variability in both dropout methods. 

Visual examples of uncertainty maps for MC-Signal and Frequency Dropout are shown in (\cref{ue_map}).

\section{Discussion and Conclusion}
\label{sec:conclusions}
This paper explores two Monte Carlo (MC) Dropout approaches for medical imaging segmentation: traditional (Signal) Dropout, applied directly to feature maps, and Frequency Dropout, operating in the frequency domain. MC simulations generated uncertainty estimates to identify segmentation errors in tasks challenging state-of-the-art models. Our results show that Frequency Dropout produced better-calibrated uncertainty estimates (\cref{best_ece}) while maintaining stable segmentation performance compared to Signal Dropout (\cref{table1}).

Semantic segmentation requires processing spatial information while preserving dependencies between related elements. Traditional Dropout, applied independently to feature map elements, can disrupt local correlations. In contrast, Frequency Dropout operates globally, preserving structural coherence and generating multiple variations of structural information, improving structural uncertainty estimation.

Traditional Dropout’s impulse noise effect varies with feature map intensity: near-zero intensities create subtler effects, while broader ranges amplify noise. Frequency Dropout provides consistent variation, unaffected by intensity scaling.

While both Signal and Frequency Dropout run efficiently on modern hardware,  Frequency Dropout's extra complexity could slow inference for large feature maps. This trade-off between processing time and enhanced uncertainty estimation should be weighed to determines suitability for specific tasks.

MC-Dropout has limitations; our study found that uncertainty estimates depend on layer placement, dropout rates, and MC repetitions (\cref{rates}). Optimal parameters vary across modalities and tasks, reflecting the method's task-dependent nature. Despite these challenges, MC-Dropout remains popular for its simplicity. Optimizing strategies in state-of-the-art architectures like U-Net could provide valuable insights.

\section{Compliance with ethical standards}
\label{sec:ethics}
This research study was conducted retrospectively using human subject data made available in open access by  \cite{antonelli2022medical, candemir2013lung, jaeger2013automatic}. 

\section{Acknowledgments}
\label{sec:acknowledgments}

No funding was received to conduct this study. The authors have no relevant financial or non-financial interests to disclose.



\bibliographystyle{IEEEbib}
\bibliography{refs}

\end{document}

%% file: figures/figure1.tex
\begin{figure}[t]

    \includegraphics[width = 8.5cm]{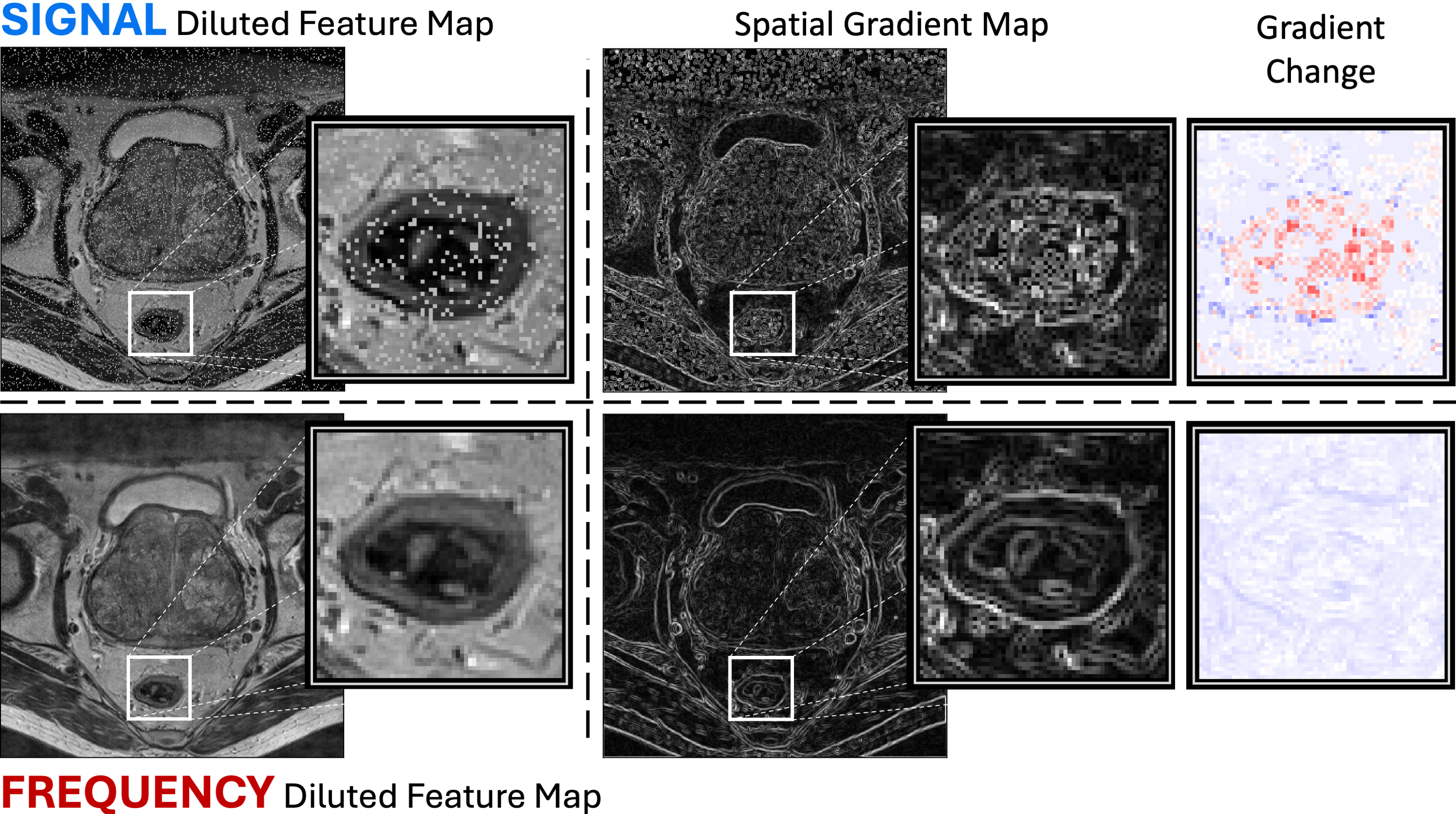}

\caption{\small Impact of Signal and Frequency Dropout on Structural Integrity. Left subplots show feature maps with traditional (Signal) Dropout ($p$ = 0.1, top) and Frequency Dropout ($p$ = 0.1, bottom). The middle subplots display Sobel gradient maps, while the right subplots illustrate structural changes by comparing pre- and post-Dropout gradients (red indicates intensified gradients, suggesting the emergence of new structures, and blue reflects diminished gradients, revealing blurring and disruption of existing structures). Signal Dropout introduces non-uniform changes in the gradient map, creating new edges while disrupting existing ones. In contrast, Frequency Dropout maintains more uniform changes across imaging features, preserving spatial dependencies. For example, gradient changes follow edges, rather than appearing irregular at the voxel level.}

\label{fig1}

\end{figure}